\documentclass[10pt,twocolumn,letterpaper]{article}

\usepackage[pagenumbers]{iccv} 

\definecolor{iccvblue}{rgb}{0.21,0.49,0.74}
\usepackage[pagebackref,breaklinks,colorlinks,allcolors=iccvblue]{hyperref}
\usepackage{multirow}
\def\paperID{4533} 
\def\confName{ICCV}
\def\confYear{2025}

\title{PCR-GS: COLMAP-Free 3D Gaussian Splatting via Pose Co-Regularizations}

\author{Yu Wei$^{1}$ \quad Jiahui Zhang$^{1}$ \quad Xiaoqin Zhang$^{2}$ \quad Ling Shao$^{3}$ \quad Shijian Lu$^{1}$ \\
\\
$^{1}$Nanyang Technological University \quad 
$^{2}$Zhejiang University of Technology \\[1mm]
$^{3}$UCAS-Terminus AI Lab, University of Chinese Academy of Sciences
}

\begin{document}

\twocolumn[{
    \renewcommand\twocolumn[1][]{#1}
    \maketitle
    \begin{center}
        \captionsetup{type=figure}
        \includegraphics[width=1.\linewidth]{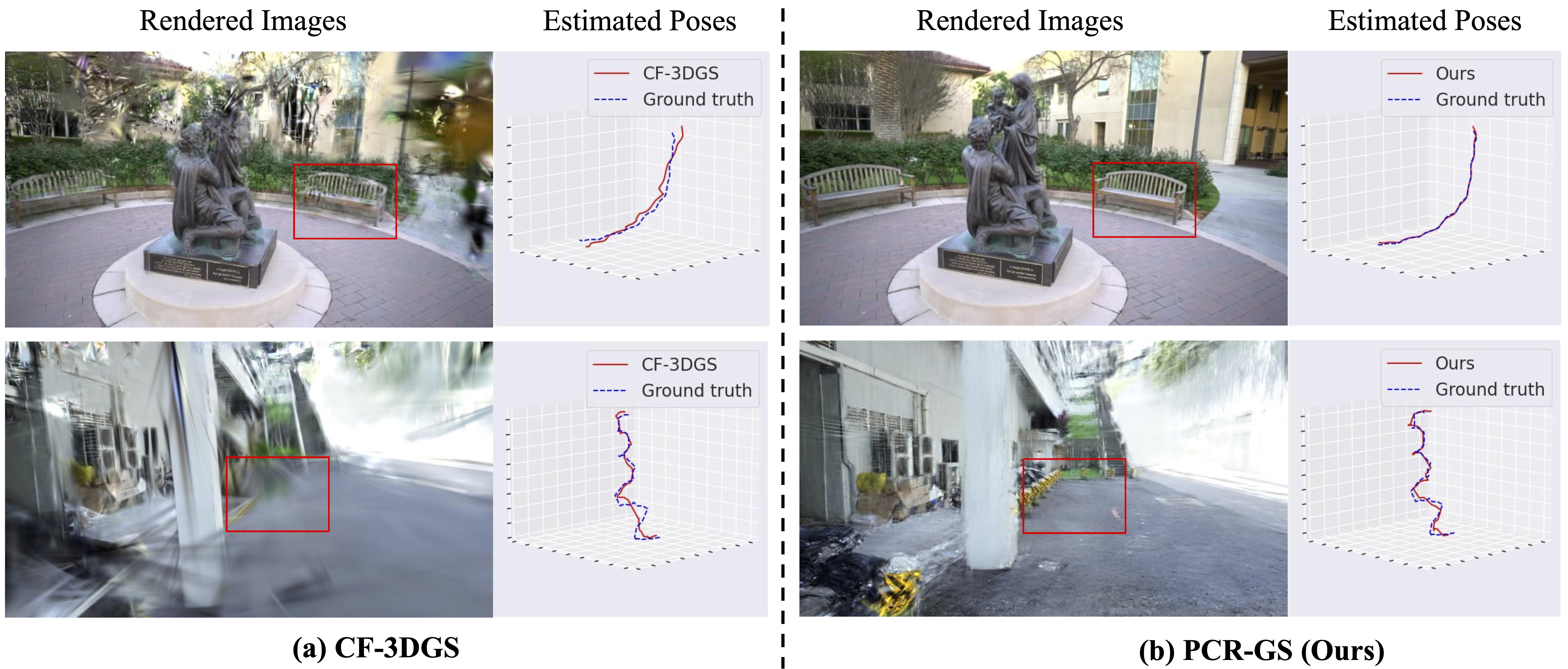}
        \captionof{figure}{
        The proposed PCR-GS can model scenes with complex camera trajectories without using any camera pose priors. It achieves more accurate camera pose estimation and more realistic novel view synthesis as compared with the state-of-the-art method CF-3DGS~\cite{fu2024colmap}. The graphs in (a) and (b) show the \textit{Rendered Images} and the \textit{Estimated Poses} by CF-3DGS and our proposed PCR-GS, respectively.
        }
        \label{teaser}
    \end{center}

}]

\maketitle
\begin{abstract}

COLMAP-free 3D Gaussian Splatting (3D-GS) has recently attracted increasing attention due to its remarkable performance in reconstructing high-quality 3D scenes from unposed images or videos. However, it often struggles to handle scenes with complex camera trajectories as featured by drastic rotation and translation across adjacent camera views, leading to degraded estimation of camera poses and further local minima in joint optimization of camera poses and 3D-GS. We propose PCR-GS, an innovative COLMAP-free 3DGS technique that achieves superior 3D scene modeling and camera pose estimation via camera pose co-regularization. PCR-GS achieves regularization from two perspectives. The first is feature reprojection regularization which extracts view-robust DINO features from adjacent camera views and aligns their semantic information for camera pose regularization. The second is wavelet-based frequency regularization which exploits discrepancy in high-frequency details to further optimize the rotation matrix in camera poses. Extensive experiments over multiple real-world scenes show that the proposed PCR-GS achieves superior pose-free 3D-GS scene modeling under dramatic changes of camera trajectories.

\end{abstract}    
\section{Introduction}
\label{sec:intro}

Photo-realistic 3D scene reconstruction and rendering has attracted increasing attention due to a wide range of applications in embodied artificial intelligence, 
virtual reality, etc. With the advent of 3D Gaussian Splatting (3D-GS)~\cite{kerbl20233d}, 3D reconstruction has made great progress by leveraging learnable 3D Gaussians to explicitly model scenes. Given a sequence of RGB images and corresponding camera poses, 3D-GS and its variants~\cite{zhou2024drivinggaussian, xiong2023sparsegs} have demonstrated superior performance in novel view synthesis. However, the impressive performance relies heavily on the availability of accurate camera poses, which are typically obtained with COLMAP~\cite{schonberger2016structure}, a Structure-from-Motion (SfM) technique that is computationally intensive and often fails for images with sparse textures or repetitive patterns.

Several studies explore pose-free 3D-GS for novel view system without using COLMAP. One representative work is COLMAP-Free 3D Gaussian Splatting (CF-3DGS)~\cite{fu2024colmap} which optimizes camera poses and 3D Gaussians sequentially as cameras move. Specifically, CF-3DGS estimates the relative camera pose between adjacent camera views and employs it to regularize camera pose estimation based on the alignment between the adjacent views. However, the accuracy of relative pose estimation is highly susceptible to complex camera trajectories, as dramatic camera movements can easily lead to limited overlaps and further compromised alignment across adjacent camera views. With inaccurate relative camera poses, joint optimization of camera poses and 3D Gaussians tends to converge to local optima as illustrated in Fig.~\ref{teaser}(a).

We propose PCR-GS, a COLMAP-free 3D-GS that co-regularizes camera poses from the perspective of feature reprojection and wavelet-based frequency regularization. Leveraging the robust DINO features against viewpoint changes~\cite{amir2021deep}, feature reprojection exploits DINO~\cite{caron2021emerging} to extract semantic features from each camera view and performs feature reprojection across adjacent camera views. 
It regularizes the estimation of relative camera poses via semantic feature alignment that minimizes the feature discrepancy between the re-projected and the target camera views. We also design an initialization strategy that initializes relative camera poses by establishing sparse feature correspondences across adjacent camera views. Such initialization facilitates the subsequent feature reprojection and mitigates the risk of local minima in camera pose estimation.

The wavelet-based frequency regularization focuses on optimizing the rotation matrix of camera poses. Specifically, rotational errors cause spatial shifts of geometries and textures, leading to loss or distortion that is well represented in high-frequency details. RGB-space regularization primarily focuses on pixel intensity changes, which is insensitive to structural shifts caused by small angular rotations. In contrast, frequency-space regularization captures rotational errors and structural shifts well with the decomposed high-frequency information. We achieve the frequency regularization with wavelet transform by decomposing images into multiple frequency bands, each having different levels of image details. As illustrated in Fig.~\ref{teaser}(b), PCR-GS achieves superior camera pose estimation and novel view synthesis without COLMAP priors.

The contribution of this work can be summarized in three aspects. 
First, we propose PCR-GS, an innovative COLMAP-free 3D Gaussian Splatting that introduces pose co-regularization and achieves superior camera pose estimation and 3D scene reconstruction under drastic camera movements across adjacent views.
Second, we leverage the robustness of DINO features against viewpoint changes and propose feature reprojection regularization which regularizes relative camera poses by reprojecting extracted DINO features between adjacent views.
Third, we design a wavelet-based frequency regularization to ensure the accurate estimation of the rotation matrix of camera poses. It can extract multi-level high-frequency components to leverage various levels of image detail to amplify the error of the rotation matrix, enabling the model to focus on the accurate prediction of the rotation matrix of camera poses.

\begin{figure*}[ht]
\begin{center}
\includegraphics[width=1\linewidth]{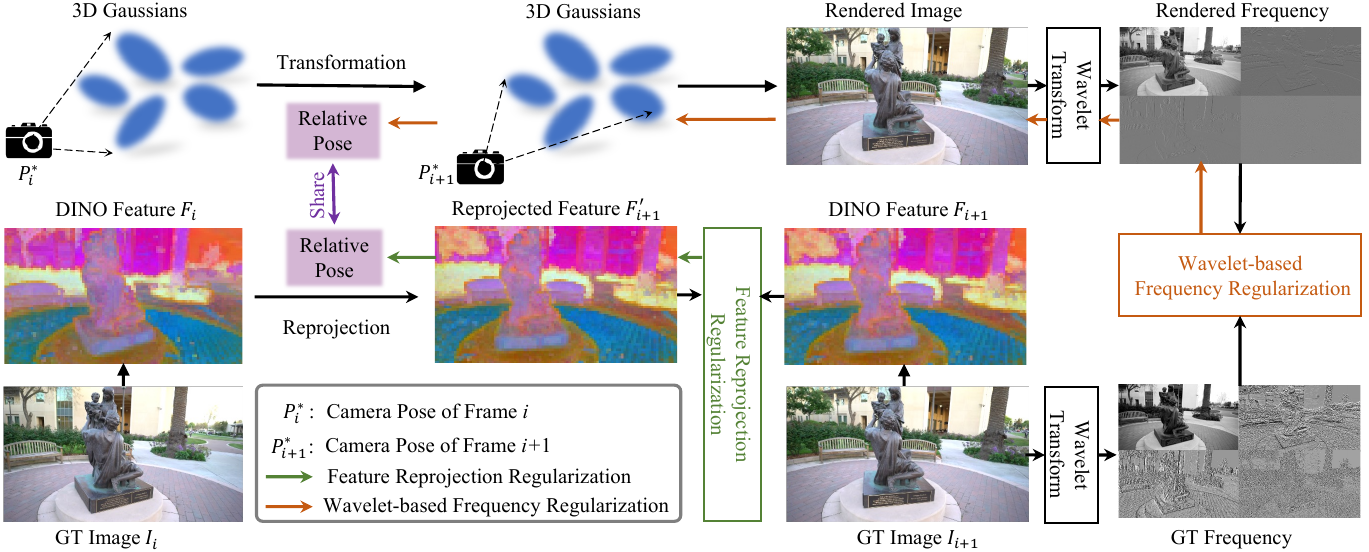}
\end{center}
\caption{
\textbf{
The framework of the proposed PCR-GS.} 
PCR-GS performs feature reprojection regularization 
and wavelet-based frequency regularization 
concurrently to optimize relative camera poses. For the feature reprojection regularization, we adopt DINO to extract semantic feature maps $F_i$ and $F_{i+1}$ of adjacent frames $I_i$ and $I_{i+1}$, and optimize the relative camera pose by minimizing the discrepancies between the reprojected feature $F'_{i+1}$ and $F_{i+1}$. For the wavelet-based frequency regularization, we employ the relative pose to transform the camera pose from $P^*_{i}$ to $P^*_{i+1}$ to render an image under $P^*_{i+1}$, and then apply wavelet transform to the rendered image and the ground-truth image to optimize the relative camera pose via frequency-space regularization. 
}
\label{overview}
\end{figure*}
\section{Related Work}
\label{sec:related}

\subsection{3D Scene Representation and Rendering}

Traditional 3D representations, such as meshes \cite{riegler2020free, riegler2021stable}, multi-plane images \cite{zhou2018stereo, tucker2020single}, and point clouds \cite{xu2022point,zhang2022differentiable}, are widely adopted to depict 3D geometry explicitly. 

Recently, Neural Radiance Fields (NeRFs) \cite{mildenhall2021nerf} have demonstrated an exceptional capability for high-quality 3D representation. They implicitly employ MLPs to reconstruct 3D scenes from multi-view posed images and achieve photorealistic novel view synthesis by differential volume rendering. Due to their nature of multi-view consistency, many NeRF variants have been developed to handle various new challenges, such as dynamic scenes \cite{pumarola2021d,liu2024dynvideo,luthra2024deblur}, few-shot modeling \cite{kwak2023geconerf,jain2021putting,wang2023sparsenerf}, model generalization \cite{ye2023featurenerf,lee2023dense,hu2023sherf} and large-scale scenes \cite{zhenxing2022switch,tancik2022block}.

However, NeRF often suffers from time-consuming training and rendering as volume rendering requires dense sampling with ray marching. Several methods \cite{muller2022instant,hu2022efficientnerf,garbin2021fastnerf} have been proposed to accelerate the training and rendering of NeRF. For instance, Muller et al. \cite{muller2022instant} introduce multi-resolution hash encoding that maps spatial coordinates to feature vectors, greatly shortening the training and rendering times. Nonetheless, these acceleration methods often sacrifice rendering quality for fast training. Kerbl et al. recently propose 3D Gaussian Splatting (3D-GS) \cite{kerbl20233d}, which achieves real-time, high-quality rendering with explicit 3D Gaussians and efficient splatting. Several 3D-GS variants have been proposed to address various new challenges, such as dynamic scenes \cite{guo2024motion,zhou2024drivinggaussian,huang2024textit}, sparse-view settings \cite{xiong2023sparsegs,zhang2025cor,zhu2025fsgs}, and SLAM \cite{li2025sgs,zhu2024semgauss,deng2024compact}. However, the impressive performance of 3D-GS relies heavily on pre-computed camera parameters which are often obtained with COLMAP. This has motivated the development of COLMAP-free algorithms that aim to remove this dependency and enhance flexibility in 3D scene reconstruction.

\subsection{Pose-Free Neural Fields}  

Recently, several studies~\cite{wang2021nerf, lin2021barf, meng2021gnerf, zhang2022vmrf, bian2023nope, fu2024colmap} investigate how to train NeRF and 3D-GS without camera pose priors. For example, NeRFmm~\cite{wang2021nerf} introduces learnable camera parameters and jointly optimizes them with NeRF. BARF~\cite{lin2021barf} introduces a progressive joint optimization approach for both camera poses and NeRF by employing a coarse-to-fine positional encoding strategy. Zhang et al.~\cite{zhang2022vmrf} propose VMRF that leverages unbalanced optimal transport to align features between rendered images and ground truth, thereby optimizing relative camera poses and NeRF scene representations. Nope-NeRF~\cite{bian2023nope} constrains relative poses by using undistorted depth prior. CF-3DGS~\cite{fu2024colmap}, the pioneer study on pose-free 3DGS, estimates relative camera poses of adjacent frames and achieves progressive pose refinement and 3DGS-based scene modeling. However, most of these methods suffer from significant performance degradation while handling scenes with drastic camera rotations and translations across adjacent views. The proposed PCR-GS introduces novel pose co-regularization, achieving superior rendering quality and pose estimation accuracy under complex camera trajectories.
\section{Method}
\label{sec:method}

This section presents the proposed PCR-GS that achieves pose-free and COLMAP-Free 3D-GS under complex camera trajectories. We first provide a brief review of the original 3D-GS and the adopted baseline CF-3DGS~\cite{fu2024colmap} in Sec.~\ref{subsec:preliminay}. Sec.~\ref{subsec:feature} then describes the proposed feature reprojection regularization that exploits the robust DINO features to estimate relative camera poses against drastic camera view changes, as well as an initialization strategy that extracts feature correspondences across adjacent views. Sec.~\ref{subsec:frequency} presents wavelet-based frequency regularization that further optimizes the rotation matrix of relative camera poses. Fig.~\ref{overview} shows the overview of PCR-GS.

\subsection{Preliminaries}
\label{subsec:preliminay}

\paragraph{3D-GS.} 
3D-GS~\cite{kerbl20233d} explicitly represents scenes with 3D Gaussians, offering advantages in differentiability and scalability. Initialized from COLMAP-generated sparse point cloud, each Gaussian is defined by a center $\mu$, covariance matrix $\Sigma$, opacity $\alpha$, and spherical harmonics (SH) coefficients $c$. 
The covariance matrix is decomposed into a scaling and rotation matrix for 
differentiable optimization.

Rendering projects 3D Gaussians onto a 2D image plane based on the camera pose. Pixel colors are computed via $\alpha$-blending of $N$ overlapping Gaussians:

\begin{equation}
    C = \sum_{i=1}^{N} c_i \alpha_i \prod_{j=1}^{i-1} (1 - \alpha_j),
\end{equation}

where $c_i$ and $\alpha_i$ are obtained by multiplying $\Sigma_i$ of the $i$-th 2D Gaussian with SH coefficients and opacity.


\paragraph{COLMAP-Free 3DGS (CF-3DGS).} 

CF-3DGS~\cite{fu2024colmap} introduces a set of 3D Gaussians to predict camera poses between nearby frames. Specifically, CF-3DGS initializes a set of 3D Gaussians $G_i$ based on a single frame $I_i$. It extracts a point cloud from the mono-depth map generated by DPT~\cite{ranftl2021vision} and sets the camera pose of frame $I_i$ as an identity matrix. $G_i$ is learned by minimizing the photometric loss between the rendered image and $I_i$:

\begin{equation}
    G_i^* = \arg \min_{c_i, \Sigma_i, \alpha_i} \mathcal{L}_{\text{rgb}}(\mathcal{R}(G_i), I_i),
\end{equation}

where $\mathcal{R}$ represents the 3D-GS rendering process for $G_i$, and $c_i$, $\Sigma_i$, and $\alpha_i$ are Gaussian parameters.

To estimate relative camera poses of adjacent frames, CF-3DGS transforms the pre-trained 3D Gaussians $G_i$ into frame $(i+1)$ using a transformation matrix $T_i$, defined as $T_i = P_i^{-1} P_{i+1}$, where $P_i$ is the camera pose of frame $i$. $T_i$ is optimized by minimizing the photometric loss between the rendered image from transformed 3D Gaussians in frame $(i+1)$ and the corresponding ground truth $I_{i+1}$:

\begin{equation}
    T_i^* = \arg \min_{T_i} \mathcal{L}_{\text{rgb}}(\mathcal{R}(T_i \circ G_i), I_{i+1}),
\end{equation}

where $\circ$ denotes the transformation from $G_i$ to $G_{i+1}$.
The photometric loss combines an L1 loss with D-SSIM:

\begin{equation}
    \mathcal{L}_{\text{rgb}} = (1-\lambda)||I_i - \hat{I_i}|| + \lambda \mathcal{L}_{\text{D-SSIM}},
\end{equation}

where $\lambda=0.2$ and $\hat{I_i}$ is the rendered image of frame $i$.

\begin{table*}[t]
\definecolor{red}{rgb}{1,0.6,0.6}
\definecolor{orange}{rgb}{1,0.8,0.6}
\definecolor{yellow}{rgb}{1,1,0.6}
        \renewcommand\arraystretch{1.0}
	\renewcommand\tabcolsep{1.0pt}
        \resizebox{1.0\textwidth}{!}{
		\begin{tabular}{l|ccc|ccc|ccc|ccc|ccc}
                \hline
			Methods   & \multicolumn{3}{c|}{CF-3DGS} & \multicolumn{3}{c|}{Nope-NeRF} &  \multicolumn{3}{c|}{Barf}  &  \multicolumn{3}{c|}{NeRFmm} & \multicolumn{3}{c}{PCR-GS (Ours)}\\
                \hline
			Scenes
			&  PSNR$^\uparrow$ & SSIM$^\uparrow$    & LPIPS$^\downarrow$  
			& PSNR$^\uparrow$  & SSIM$^\uparrow$   & LPIPS$^\downarrow$  
			& PSNR$^\uparrow$  & SSIM$^\uparrow$   & LPIPS$^\downarrow$
                & PSNR$^\uparrow$ & SSIM$^\uparrow$   & LPIPS$^\downarrow$
                & PSNR$^\uparrow$ & SSIM$^\uparrow$    & LPIPS$^\downarrow$
                \\
			\hline 
			Church   & 27.71 & 0.88  & {0.13} & 23.32 & 0.65 &0.47 & 13.74 & 0.37 & 0.70 &16.94& 0.42&0.61 & \textbf{27.90} & \textbf{0.89} & \textbf{0.12} \\
			Barn   & 23.49 & 0.71 & 0.22 & 23.34 & 0.60 & 0.51 &15.69 & 0.52 & 0.60 & 17.51& 0.53&0.58 & \textbf{28.57} & \textbf{0.84} & \textbf{0.13} \\ 
			Museum  & 18.43 & 0.54 & 0.37 &  \textbf{22.42} & 0.55 & 0.51 & 14.75 & 0.39 & 0.65 &11.43 & 0.22 & 0.74 & 21.56 & \textbf{0.66} & \textbf{0.25} \\
                
                Family  & 18.03  & 0.57 & 0.37 & 22.72 &0.59 & 0.54  & 13.94 & 0.43 & 0.70 & 14.11 &0.37 & 0.67& \textbf{25.45} & \textbf{0.82} & \textbf{0.16}\\
                Horse  & 18.34  & 0.64 & 0.32 &  21.74 & 0.66 & 0.44 & 13.43&0.54 & 0.63 & 13.31 &0.46 &0.62& \textbf{24.20} & \textbf{0.79} & \textbf{0.17}  \\
                Ballroom  &   17.05& 0.48 & 0.39 & 18.84 & 0.46 & 0.58 &12.55 & 0.31 & 0.76 & 12.43 & 0.24&0.67& \textbf {19.70} & \textbf {0.60}& \textbf{0.30}\\
                Francis  & 16.23  & 0.50 & 0.49 &  21.72 & 0.59 & 0.59  & 15.38 & 0.52 & 0.68 & 13.19& 0.38& 0.69 & \textbf{21.91} & \textbf{0.65} & \textbf{0.34}\\
                Ignatius & 19.07  & 0.49 & 0.35 & \textbf{21.49} & 0.47 & 0.58  & 12.93 & 0.31 & 0.79 &13.95&0.33&0.66 & 20.15 & \textbf{0.60} & \textbf{0.33}\\
                \hline
                Mean   & 19.79 & 0.60 & 0.33 & 21.95 & 0.57 & 0.52 & 14.05 & 0.42& 0.69& 14.10& 0.36& 0.66 & \textbf{23.68} & \textbf{0.73} & \textbf{0.23}\\
                \hline
			
		\end{tabular}
            }
	\caption{\textbf{Quantitative comparisons of novel view synthesis on Tanks\&Temples~\cite{knapitsch2017tanks}.} 
    Each baseline method is trained with its public code under the original settings and evaluated with the same evaluation protocol. The best score of all the results is in bold. Note the Tanks\&Temples here has more drastic camera motions as detailed in the Dataset part and appendix.}
 \label{quantitative_result_nvs}
\end{table*}

\begin{figure}[t]
\begin{center}
\includegraphics[width=1\linewidth]{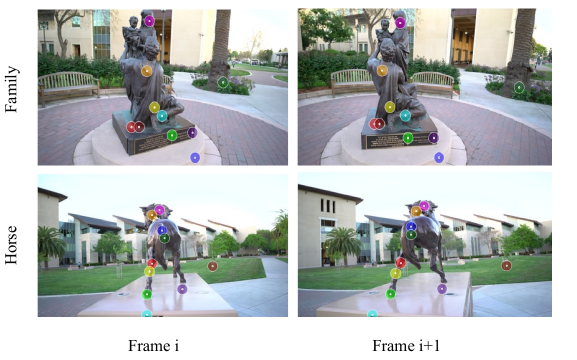}
\end{center}
\caption{
\textbf{Visualization of robust DINO features against drastic camera motion.} Point correspondences as searched by DINO feature maps remain stable under drastic camera pose changes.
}
\label{correspondence_result}
\end{figure}

While dealing with scenes with complex camera trajectories, the relative camera pose may experience dramatic changes due to large camera movement. Under this circumstance, regularizing camera poses over RGB images becomes susceptible due to limited overlaps across adjacent frames. In comparison, DINO features are much more stable under drastically view changess \cite{amir2021deep} as illustrated in Fig.~\ref{correspondence_result} 
We therefore align the robust DINO-based semantic features across adjacent views to optimize the camera pose with drastic rotations and translations. Specifically, with DINO features extracted from every frame, we design a feature reprojection regularization technique that optimizes relative camera pose as illustrated in Fig. \ref{overview}.

\subsection{Feature Reprojection Regularization}
\label{subsec:feature}

The pre-trained \(G_i^{*}\) can be rendered into RGB images and depth maps under the camera pose \(P_i\), from which we can obtain the depth value of each pixel on \(I_i\): 
\begin{equation}
D = \sum_{i=1}^{N} d_i \alpha_i \prod_{j=1}^{i-1} (1 - \alpha_j),
\end{equation}
where $D$ is the depth value of each pixel, and \(d_i\) denotes the distance from the \(i_{th}\) overlapping Gaussian to the camera.

We regularize the relative camera pose by aligning DINO features across two adjacent views. Specifically, we first reproject 2D pixels of \(I_i\) into 3D camera coordinate \(P_i\) and then transform them into the adjacent camera coordinate \(P_{i+1}\) with the transformation matrix \(T_i\). The transformed 3D points are then projected back to 2D image \(I_{i+1}\). By sampling features from two adjacent feature maps based on their original and reprojected positions, we minimize the discrepancy between features from $I_i$ and the reprojected features from $I_{i+1}$. During the projection and reprojection, we use the camera intrinsic matrix \( \mathbf{K} \) to transform points across 3D camera coordinate and 2D image coordinate.

\begin{equation}
    T_i^* = \arg \min_{T_i} \mathcal{L}_{\text{feat}}(F_i\langle P_i \rangle , F_{i+1} \left\langle \mathbf{K}  P_i T_i \right\rangle ),
\end{equation}
where \(F_i\) is the DINO feature map extracted from \(I_i\) and   
\( \langle \cdot \rangle \) represents a sampling operation on feature map. The feature loss is computed by L2 loss between feature map and reprojected feature map.
\begin{equation}
    \mathcal{L_{\text{feat}}} = ||F_i\langle P_i \rangle - F_{i+1} \left\langle \mathbf{K}  P_i T_i \right\rangle||_{2}.
\end{equation}

In addition, poor initialization of camera poses can trap the optimization in local optima, especially when the camera poses experience dramatic changes. We establish sparse correspondences between adjacent frames by searching the DINO features as represented by $CP_i$:
\begin{equation}
    CP_i = \left\{ (p, q) \mid p \in I_i, \, q \in I_{i+1}\right\},
\end{equation}
where \(p\) and \(q\) denote the positions of correspondent key points on the respective image. 
We then utilize a saliency map derived from DINO feature map to construct a foreground mask, filtering out background pixels and selecting candidate key points on the feature map. Next, we compute the similarity between these key feature points across frames with the Best Buddies algorithm \cite{dekel2015best} and determine correspondences by the best-matched point pairs. Since such derived correspondences often contain noises, we randomly select \(N_s = 20\) sparse correspondences to optimize the relative camera pose and apply the optimization to initialize the camera pose as follows: 
\begin{equation}
    T_i^* = \arg \min_{T_i} \sum_{s=1}^{N_s} ||\mathbf{K} p_s T_i -  \mathbf{K}q_s||.
\end{equation}

Finally, we use the optimized relative camera pose instead of the Identity matrix for initialization which improves the camera pose estimation clearly.

\subsection{Wavelet-based Frequency Regularization}
\label{subsec:frequency}


The camera pose is defined by a rotation matrix \(R \in \mathfrak{so}(3) \) and a translation vector \(t \in \mathbb{R}^3 \). Our empirical studies show that the rotation matrix is much more challenging to optimize, where a small error in \(R \in \mathfrak{so}(3) \) often leads to shifts of geometric structures such as edges and textures. Regularization in the RGB space primarily focuses on pixel intensity which is not sensitive to the structural changes due to camera rotations. We therefore propose a frequency regularization approach to further optimize the rotation matrix of camera poses. Specifically, we adopt wavelet transformation that decomposes an RGB image into multiple frequency components along horizontal and vertical directions. Unlike other frequency transformations, wavelet transform provides rich high-frequency details that retain their original spatial location within the image and also highlight subtle rotation errors for optimization.

Specifically, the wavelet transform decomposes an image \( I(u, v) \) into four components: \( LL \), \( LH \), \( HL \), and \( HH \), representing low-frequency component, horizontal high-frequency component, vertical high-frequency component and diagonal high-frequency component, respectively.
\begin{align}
    LL(u, v) &= \sum_{m} \sum_{n} I(m, n) \cdot h(u - m) \cdot h(v - n), \nonumber\\
    LH(u, v) &= \sum_{m} \sum_{n} I(m, n) \cdot h(u - m) \cdot g(v - n), \nonumber\\
    HL(u, v) &= \sum_{m} \sum_{n} I(m, n) \cdot g(u - m) \cdot h(v - n), \nonumber\\
    HH(u, v) &= \sum_{m} \sum_{n} I(m, n) \cdot g(u - m) \cdot g(v - n),
\end{align}
where \( I(m, n) \) is the pixel value at position \( (m, n) \) in the input image \(I\), and \( h(\cdot) \) and \( g(\cdot) \) respectively represent low-pass filter, and high-pass filter. 

The discrepancies of each component (denoted as \(d\)) between rendered image \(\hat{I}\) and ground truth \(I\) can be obtained with the Euclidean metric as follows: 
\begin{equation}
    d = 
\sum_{x \in \{LL,LH, HL, HH\}} w_x \left\| W_x(I_t) - W_x\left(\hat{I}_t\right) \right\|,
\label{eq9}
\end{equation}
where \(W_x\) represents the wavelet transformation that extract component \(x\) from an image and the \(w_x\) denotes the weight for each wavelet component.


Direct optimization of high-frequency components often introduces noise and complicates the network training. We address this issue by introducing an annealing strategy that gradually minimizes the discrepancy from low frequency to high frequency and accordingly achieves progressive pose regularization. Specifically, the annealing strategy regularizes low-frequency information first and then progressively increases the weight of high-frequency information in the loss computation. Such progressive frequency regularization \(\mathcal{L}_{freq}\) can be formulated as follows:
\begin{equation}
    \mathcal{L}_{freq} =
\begin{cases}
d_{LL} & 0 < n \leq n_0, \\
(1 - w_h) d_{LL} + w_h d_{H}  & n_0 < n \leq n_1, \\
d_{H}  & n > n_1,
\end{cases}
\end{equation}
where \(d_H\) denotes the discrepancies of all high-frequency components \(LH, HL, HH\) as mentioned in Eq.~\ref{eq9}. \(w_h\) denotes the weight for the high-frequency components, computed as follows:
\begin{equation}
    w_h(n) = \frac{n - n_0}{n_1 - n_0},
\end{equation}
where \(n_0 = 100\) and \(n_1 = 200\) represent the iterations for introducing high-frequency components and deactivating low-frequency components, respectively.

\subsection{Overall Training Pipeline}
\label{subsec:pipeline}
With all defined loss terms, the overall training objective can be formulated as follows:
\begin{equation}
\mathcal{L} = \lambda_0 \mathcal{L}_{\text{rgb}} + \lambda_1 \mathcal{L}_{\text{feat}} + \lambda_2 \mathcal{L}_{\text{freq}} ,
\end{equation}
where \( \lambda_0, \lambda_1\) and \(\lambda_2 \) are the weights for each loss term, setting to 0.6, 0.2 and 0.2, respectively. 

\begin{table*}[t]
\definecolor{red}{rgb}{1,0.6,0.6}
\definecolor{orange}{rgb}{1,0.8,0.6}
\definecolor{yellow}{rgb}{1,1,0.6}
        \renewcommand\arraystretch{1.0}
	\renewcommand\tabcolsep{1.0pt}
        \resizebox{1.0\textwidth}{!}{
		\begin{tabular}{l|ccc|ccc|ccc|ccc|ccc}
                \hline
			Methods   & \multicolumn{3}{c|}{CF-3DGS} & \multicolumn{3}{c|}{Nope-NeRF} &  \multicolumn{3}{c|}{Barf}  &  \multicolumn{3}{c|}{NeRFmm}& \multicolumn{3}{c}{PCR-GS (Ours)}\\
                \hline
			Scenes
			&  \(RPE_t\)$^\downarrow$ & \(RPE_r\)$^\downarrow$    & ATE$^\downarrow$  
			& \(RPE_t\)$^\downarrow$ & \(RPE_r\)$^\downarrow$    & ATE$^\downarrow$ 
			& \(RPE_t\)$^\downarrow$ & \(RPE_r\)$^\downarrow$    & ATE$^\downarrow$
                & \(RPE_t\)$^\downarrow$ & \(RPE_r\)$^\downarrow$    & ATE$^\downarrow$
                & \(RPE_t\)$^\downarrow$ & \(RPE_r\)$^\downarrow$    & ATE$^\downarrow$
                \\
			\hline 
			Church   & 0.052 & 0.079  & {0.008} & 1.426 &0.246 & 0.303 &  3.626 &1.079 & 0.579  & 3.599 &0.647 & 0.586 & \textbf{0.049} & \textbf{0.077} & \textbf{0.007}\\
			Barn   & 0.307 & 0.144 & 0.016 & 2.557 &0.723 & 0.259  & 2.968 &0.312 & 0.097  & 4.785 &0.586 & 0.141 & \textbf{0.114} & \textbf{0.103} & \textbf{0.010} \\ 
			Museum   & 0.116 & 0.606 & 0.021 &  3.134 &0.897 & 0.444 &  9.240 &2.192 & 0.459  & 9.732 &2.108 & 0.530 & \textbf{0.074} & \textbf{0.397} & \textbf{0.014} \\
                
                Family  & 0.366  & 0.618 & 0.018 & 4.086 &0.274 & 0.404  &  8.982 &2.288 & 0.229  &  8.496 &2.112 & 0.329 & \textbf{0.093} & \textbf{0.210} & \textbf{0.005}\\
                Horse  & 0.080  & 0.186 & \textbf{0.005} &  1.982 &0.474 & 0.181 & 6.540 &1.570 & 0.564 &  6.201 &1.162 & 0.535 & \textbf{0.061} & \textbf{0.126} & \textbf{0.005} \\
                Ballroom  &  0.517 &1.392 & 0.021  &  4.530 &0.985 & 0.560 & 13.595 &4.375 & 0.472 &  15.436 &4.796 & 0.515 & \textbf {0.250} & \textbf {0.854}& \textbf{0.010}\\
                Francis  & 0.216 &0.971 & 0.012 & 6.810 &2.013 & 0.531 &  11.802 &3.390 & 0.507 & 11.696 &2.200 & 0.538  & \textbf{0.197} & \textbf{0.891} & \textbf{0.010}\\
                Ignatius  & 0.032 & 0.164 & \textbf{0.006} & 3.630 &0.397 & 0.544 & 4.375 &1.761 & 0.579  & 6.150 &1.992 & 0.395& \textbf{0.030} & \textbf{0.145} & \textbf{0.006}\\
                \hline
                Mean   & 0.211 & 0.520 & 0.013 & 3.519 & 0.751 & 0.403 & 7.641 & 2.121 & 0.436 & 8.261 & 1.950 & 0.446 & \textbf{0.109} & \textbf{0.350} & \textbf{0.008}\\
                \hline
			
		\end{tabular}
            }
	\caption{\textbf{Quantitative comparisons of pose estimation on Tanks\&Temples~\cite{knapitsch2017tanks}.} 
    Each baseline method is trained with its public code under the original settings and evaluated with the same evaluation protocol. The best score of all the results is in bold. Note the Tanks\&Temples here has more drastic camera motions as detailed in the Dataset part and appendix.}
 \label{quantitative_result_pose}
\end{table*}

\section{Experiment}
\label{sec:experiment}

\subsection{Datasets and Evaluation Metrics}

\textbf{Datasets.} We conduct experiments over 11 real-world scenes with complex camera trajectories from the widely adopted benchmarks Tanks\&Temples~\cite{knapitsch2017tanks} (8 scenes), Free-Dataset~\cite{wang2023f2} (3 scenes).

\paragraph{Tanks\&Temples. } The Tanks\&Temples dataset used in CF-3DGS samples video frames at 20 fps on average from eight videos of different scenes. However, the resulting camera trajectories are overly smooth and lack camera motion complexity which is typically encountered in real-world scenarios. In contrast, our method targets pose-free 3DGS under challenging conditions with drastic camera motions and complex camera trajectories where pose estimation becomes significantly more difficult. CF-3DGS struggles in such settings due to its reliance on smooth camera trajectories, limiting its effectiveness in realistic environments. 
Thus, to evaluate our proposed method, we reconstruct this dataset by sampling video frames at 4 fps on average from the eight videos which produces much more complex camera trajectories and drastic camera motions. We retain the original training and testing data split strategy. Please refer to the appendix for details on video frame selection.

\paragraph{Free-Dataset. } The Free-Dataset comprises large-scale scenes with significant camera motion and complex camera trajectories but no central objects. We use 50 frames per scene for training and testing~\cite{ran2024ct}, with ground-truth poses from COLMAP. We select one frame as test data every 8 frames in each scene.


\paragraph{Evaluation Metrics. } For quantitative experiments, we adopt the standard evaluation metrics including PSNR, SSIM \cite{wang2004image}, and LPIPS \cite{zhang2018unreasonable} for evaluation of novel view synthesis. For pose estimation, we report relative pose rotation error (\(RPE_r\)), relative pose translation error (\(RPE_t\)), and Absolute Trajectory Error(ATE) \cite{sturm2012benchmark}.

\subsection{Implementation Details}

\paragraph{Implementation.} For every video frame, we extract the DINO features from the 9th layer of the DINO model \cite{amir2021deep} where the features can retain good details of the image structure. Following CF-3DGS \cite{fu2024colmap}, we employ the mono-depth map extracted by the dense prediction transformer (DPT) \cite{ranftl2021vision} to initialize the point cloud for each frame. 
Besides, the training process follows the incremental optimization strategy and configuration setting as in CF-3DGS. Before the pose co-regularization, we initialize the relative camera pose by establishing correspondences between adjacent views. The optimization iteration for the initialization stage is set as 200 and the learning rate gradually decays from \(10^{-4}\) to \(10^{-5}\). We utilize a quaternion rotation \(\mathbf{q} \in \mathfrak{so}(3)\), and translation vector \(\mathbf{t} \in \mathbb{R}^3\) to represent camera poses, enabling smooth optimization. 
All experiments are conducted on a single NVIDIA A100 40G GPU. More details are provided in the supplementary material.

\begin{figure*}[t]
\begin{center}
\includegraphics[width=1\linewidth]{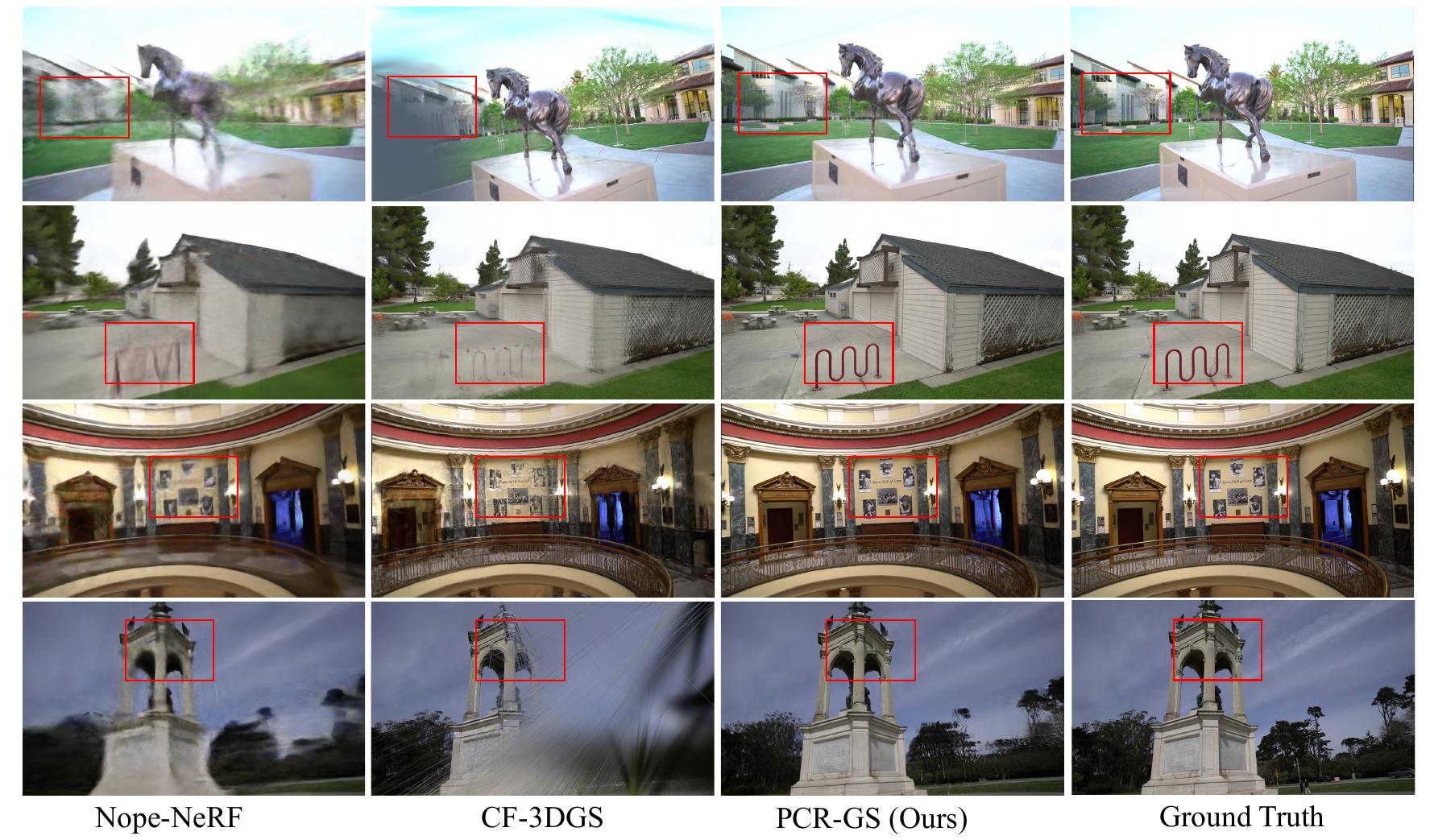}
\end{center}
\caption{
\textbf{Qualitative comparisons of PCR-GS with CF-3DGS~\cite{fu2024colmap} and Nope-NeRF~\cite{bian2023nope} in novel view synthesis (on Tanks\&Temples)}. PCR-GS achieves superior image rendering as compared with the two state-of-the-art methods.}

\label{qualitative_comparison}
\end{figure*}

\subsection{Comparisons with COLMAP-free methods}

We benchmark our method with several state-of-the-art COLMAP-free models including NeRFmm \cite{wang2021nerf}, BARF \cite{lin2021barf}, Nope-NeRF \cite{bian2023nope}, and CF-3DGS \cite{fu2024colmap}. The benchmarking focuses on two aspects: novel view synthesis and pose estimation. All the models are trained and tested over the same data and hardware for fairness.

\begin{figure}[t]
\begin{center}
\includegraphics[width=1\linewidth]{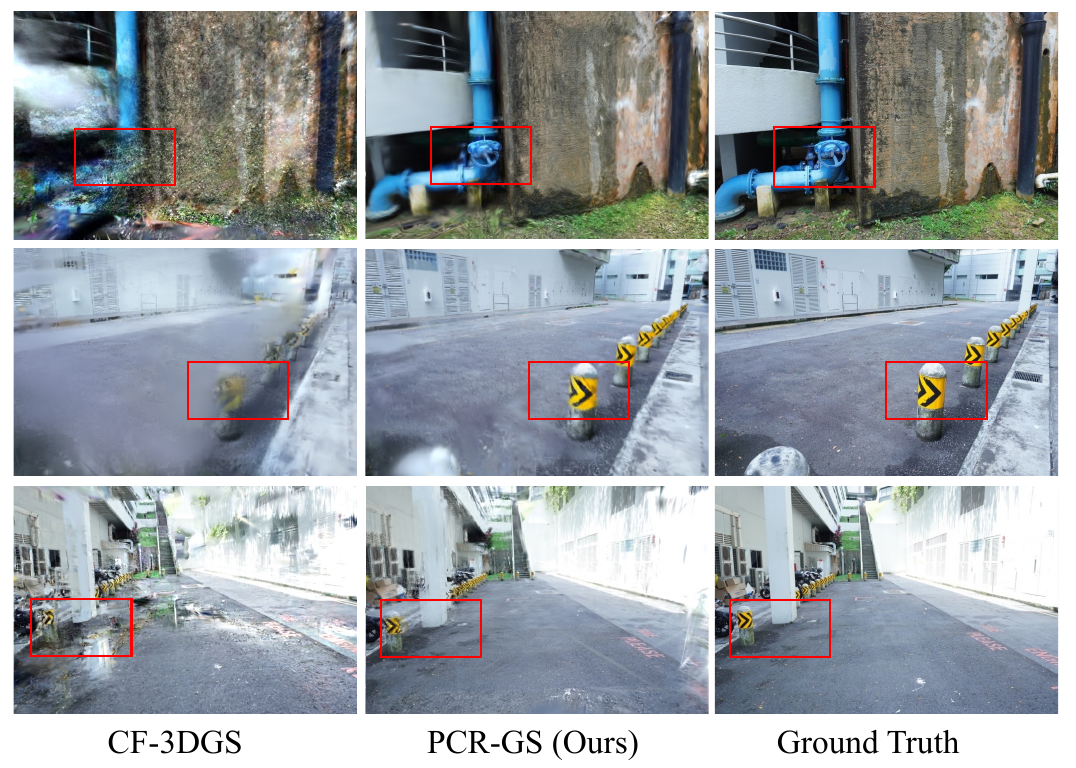}
\end{center}
\caption{
\textbf{Qualitative comparison on novel view synthesis over the Free-dataset~\cite{wang2023f2}.} The proposed PCR-GS generates better details with less artifacts consistently. 
}
\label{qualitative_comparison_2}
\end{figure}

\paragraph{Novel View Synthesis. } Since the camera pose of each test frame is unknown, we freeze the parameters of the 3DGS model trained on the training set and predict the camera pose of test frames as in CF-3DGS. We then render images on test sets and compare the quality of rendered images across different models. As Tables \ref{quantitative_result_nvs} and \ref{quantitive_comparison_nvs_2} show, our method achieves superior novel view synthesis consistently over all three metrics and 11 scenes from the two benchmarks. We also conducted qualitative experiments and obtain well-aligned experimental results. As illustrated in Fig. \ref{qualitative_comparison} and Fig.~\ref{qualitative_comparison_2}, the images synthesized by PCR-GS are noticeably clearer and produce fewer artifacts as compared with those generated by the state-of-the-art models. 
Note for Tanks\&Temples, CF-3DGS obtains much lower performance than what was presented in the original paper. The lower performance is largely attributed to the more sparsely sampled training data as detailed in the Dataset part, where we sample video frames at 4 fps (instead of 20 fps as in CF-3DGS) to have data with more dynamic camera trajectory. 

\paragraph{Pose Estimation.} We apply the Procrustes analysis as in \cite{bian2023nope} to post-process the estimated camera poses, transforming the estimated camera poses and the ground truth into a common coordinate space for comparisons. Table. \ref{quantitative_result_pose} shows experimental results on Tanks\&Temples. It can be observed that our method outperforms the CF-3DGS consistently in all scenes. We also evaluate the pose estimation on another benchmark Free-Dataset. As Table. \ref{quantitive_comparison_pose_2} shows, our method achieves better performance than CF-3DGS as well. The quantitative comparisons demonstrate the effectiveness of our proposed method in camera pose estimation.

\begin{table}[t]
    \renewcommand\arraystretch{1.35}
    \renewcommand\tabcolsep{3.0pt}
    \centering
    \resizebox{1.0\linewidth}{!}{
        \begin{tabular}{c|l|ccc|ccc}
            \hline
            & \multirow{2}{*}{Scenes} & \multicolumn{3}{c|}{CF-3DGS} & \multicolumn{3}{c}{PCR-GS (Ours)} \\
            \cline{3-8}
            &  & PSNR$^\uparrow$ & SSIM$^\uparrow$ & LPIPS$^\downarrow$ & PSNR$^\uparrow$ & SSIM$^\uparrow$ & LPIPS$^\downarrow$ \\
            \hline
            \multirow{4}{*}{\rotatebox{90}{Free-Dataset}}  
            & Pillar  & 14.47 & 0.41 & 0.61  & \textbf{17.15} & \textbf{0.49} & \textbf {0.52}\\
            & Stair  & 16.82 & 0.48& 0.48 & \textbf{20.81} & \textbf{0.60} & \textbf{0.35}\\
            & Hydrant  &14.02&0.22&0.56 & \textbf{15.37} & \textbf{0.37} & \textbf{0.52}\\
            \cline{2-8}
            & Mean &  15.00 & 0.37 & 0.55 &  \textbf {17.78} &  \textbf {0.49} &  \textbf {0.46} \\
            \hline
            
        \end{tabular}
    }
    \caption{\textbf{Quantitative comparisons on novel view synthesis over the Free-Dataset~\cite{wang2023f2}}. The best score is in bold.
	}
 \label{quantitive_comparison_nvs_2}
\end{table}

\begin{table}[t]
    \renewcommand\arraystretch{1.35}
    \renewcommand\tabcolsep{3.0pt}
    \centering
    \resizebox{1.0\linewidth}{!}{
        \begin{tabular}{c|l|ccc|ccc}
            \hline
            & \multirow{2}{*}{Scenes} & \multicolumn{3}{c|}{CF-3DGS} & \multicolumn{3}{c}{PCR-GS (Ours)} \\
            \cline{3-8}
            &  &\(RPE_t\) $^\downarrow$   & \(RPE_r\) $^\downarrow$  & ATE$^\downarrow$ & \(RPE_t\)$^\downarrow$ & \(RPE_r\)$^\downarrow$ & ATE$^\downarrow$\\
            \hline
            \multirow{4}{*}{\rotatebox{90}{Free-Dataset}}  
            & Pillar   & 0.779 & 4.482 & 0.014& \textbf{0.314} & \textbf{1.481} & \textbf {0.008}\\
            & Stair   & 1.222 & 2.205 & 0.024 & \textbf{0.654} & \textbf{0.540} & \textbf{0.013}\\
            & Hydrant  &3.609 &7.331 & 0.088 & \textbf{1.739} & \textbf{3.223} & \textbf{0.051} \\
            \cline{2-8}
            & Mean &  1.870  & 4.673 & 0.042  &\textbf {0.902} &  \textbf {1.748} &  \textbf {0.024} \\
            \hline
            
        \end{tabular}
    }
    \caption{\textbf{Quantitative comparisons on pose estimation over the Free-Dataset~\cite{wang2023f2}}. The best score is in bold.
	}
 \label{quantitive_comparison_pose_2}
\end{table}

\begin{table}[t]
\renewcommand\arraystretch{1.1}
\renewcommand\tabcolsep{1.0pt}
\begin{center}
    
\begin{tabular}{l||ccc} 
\hline
& 
\multicolumn{3}{c}{Evaluation Metrics}
\\
\cline{2-4}
\multirow{-2}{*}{Models} 
& PSNR $\uparrow$ 
& SSIM $\uparrow$
& LPIPS $\downarrow$ 
\\\hline
Base (CF-3DGS) & 18.34 & 0.64 & 0.32 \\

Base+FRR & 23.16 & 0.72 & 0.17 \\

Base+WRF(w/o high-freq) &18.40 & 0.65 & 0.32 \\

Base+WRF(w/ high-freq) & 19.31 & 0.66 & 0.28  \\

Base+FRR+WRF(w/ high-freq) & \textbf{24.20} & \textbf{0.79} & \textbf{0.17} \\

\hline
\end{tabular}
\end{center} 
\caption{
\textbf{
Ablation studies of PCR-GS on Tanks\& Temples}. We report the performance on novel view synthesis with metrics PSNR, SSIM and LPIPS. With CF-3DGS as the baseline model \textit{Base}, \textit{Base+FRR} and \textit{Base+WFR} introduce feature reprojection regularization (FRR) and wavelet-based frequency regularization (WFR), respectively. High-freq denotes the high-frequency component in wavelet-based frequency regularization. \textit{Base+FRR+WFR} (i.e. PCR-GS) further introduces WFR on top of \textit{Base+FRR} to optimize the rotation matrix in camera poses. The best results are highlighted in bold.
}
\label{ablation}
\end{table}

\subsection{Ablation Study}

We conduct ablation experiments to examine how our proposed feature reprojection regularization and wavelet-based frequency regularization contribute to  scene reconstruction. 

\paragraph{Feature Reprojection Regularization}

We first examine how our proposed feature reprojection regularization affects PSNR, SSIM, and LPIPS in Gaussian Splatting. We adopt CF-3DGS as the baseline \textit{Base}, which simply uses RGB regularization to constrain camera pose estimation. On top of the \textit{Base}, we train a new model \textit{Base+FRR} that incorporates our proposed feature reprojection regularization. As Table.\ref{ablation} shows, the \textit{Base+FRR} outperforms the \textit{Base} clearly in PSNR, SSIM and LPIPS, demonstrating the effectiveness of the proposed feature reprojection regularization. 

\paragraph{Wavelet-based Frequency Regularization}


We train two independent models that incorporate WFR on top of the \textit{Base}. One model utilizes the high-frequency component, while the other does not. On top of the \textit{Base+FRR}, we train a new model \textit{Base+FRR+WFR} (i.e. complete PCR-GS) that further incorporates the proposed wavelet-based frequency regularization to evaluate how it contributes to the 3D scene reconstruction and novel view synthesis. Table \ref{ablation} shows experimental results. We can observe that the proposed wavelet-based frequency regularization further improves the novel view synthesis effectively, with the high-frequency component playing an important role.

\section{Conclusion}
This paper presents PCR-GS, an innovative COLMAP-free 3DGS method that leverages pose co-regularization to achieve superior rendering quality and pose estimation accuracy while handling scenes with complex camera trajectories. Specifically, we design the feature reprojection regularization that extracts view-robust DINO features and aligns the semantic information between adjacent frames to regularize the relative camera poses. In addition, we design wavelet-based frequency regularization to regularize the rotation matrix of camera poses by minimizing discrepancies in high-frequency details.
Extensive experiments demonstrate the effectiveness of our method on large-scale scenes with complex camera trajectories.
\section{Acknowledgements}
\label{sec:ack}
This project is funded by the Ministry of Education Singapore, under the Tier-1 project with the project number RT18/22.

{
    \small
    \bibliographystyle{ieeenat_fullname}
    \bibliography{main}
}

\end{document}